%% file: main.tex
\documentclass[10pt,twocolumn,letterpaper]{article}

\usepackage{etoolbox,lipsum}

\usepackage[pagenumbers]{cvpr} 

\input{preamble}

\definecolor{cvprblue}{rgb}{0.21,0.49,0.74}
\usepackage[pagebackref,breaklinks,colorlinks,citecolor=cvprblue]{hyperref}

\def\paperID{13400} 
\def\confName{CVPR}
\def\confYear{2024}

\title{Action-based image editing guided by human instructions}

\author{Maria Mihaela Trusca$^1$ \qquad Mingxiao Li$^2$ \qquad Marie-Francine Moens$^2$\\
$^1$Faculty of Arts, KU Leuven, Leuven, Belgium \\
$^2$Department of Computer Science, KU Leuven, Leuven, Belgium \\
{\tt\small {\{mariamihaela.trusca, mingxiao.li, sien.moens\}}@kuleuven.be}}

\begin{document}

\twocolumn[{%
\renewcommand\twocolumn[1][]{#1}%
\maketitle
    \begin{center}
    \centering
    \captionsetup{type=figure}
    \includegraphics[width=\textwidth]{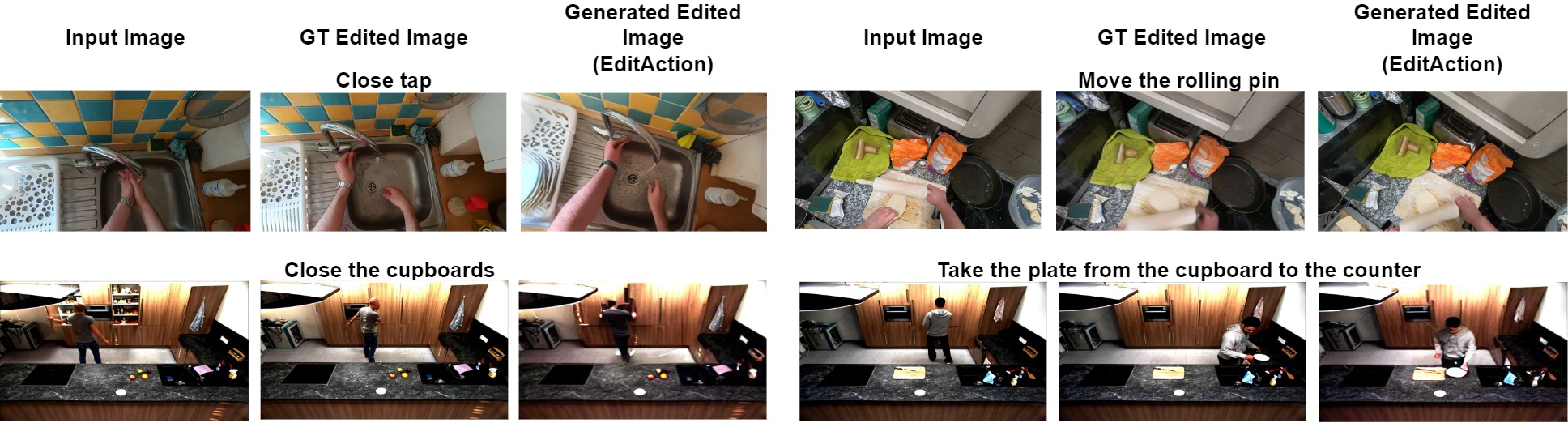}
    \caption{Given a text instruction describing an action, we propose \textbf{EditAction}, a method for editing the position and posture of objects in an input image to visually depict the action while preserving the appearance of the objects involved in the action. 
    }\label{fig:first_figure}
    \end{center}
}]

\input{sec/0_abstract}    
\input{sec/1_intro}

\input{sec/2_formatting}
\input{sec/3_finalcopy}
{
    \small
    \bibliographystyle{ieeenat_fullname}
    \bibliography{main}
}

\clearpage
\section{Appendix}
\subsection{LC and HC datasets}\label{sec:datasets}

\textbf{LC dataset. }While 66.49\% of LC text instructions specify a starting point for the action, only 34.11\% indicate the ending point. In 29.08\% of cases, the text instruction specifies only the action verb and the target object of the activity. After extracting video frames, the LC dataset comprises 13766 image pairs that show 61 different actions. 

\textbf{HC dataset. }In the case of HC, most text instructions specify only the action verb and the object, with only 18.90\% of the instructions providing an ending point for the activity, while the starting point is always omitted. As observed on the left side of Figure~\ref{fig:datasets}, the ending point of the activity is usually specified for long-distance actions that require moving the object to a location outside the central visual field (e.g., the ending point "bin" in the lower-left example presented in Figure~\ref{fig:datasets}). After processing the videos, the HC dataset comprises 37454 image pairs that depict 260 different actions.

\subsection{The selection of $\lambda_1$ and $\lambda_2$}\label{sec:editaction_hyperparamets}
\begin{figure}[b]
    \centering
    \includegraphics[width=8cm]{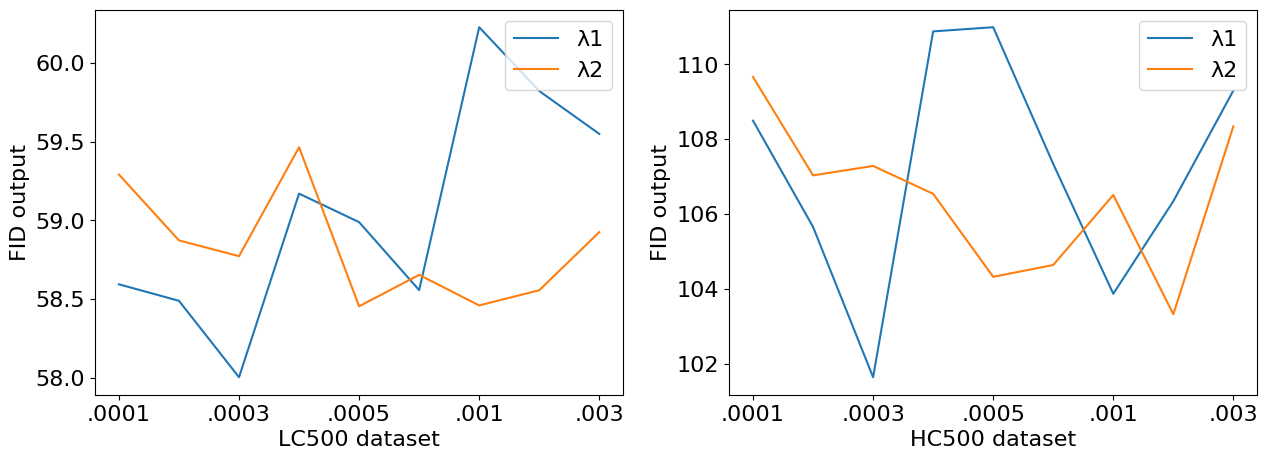}
    \caption{The selection of $\lambda_1$ and $\lambda_2$ hyperparameters based on the $FID_{output}$ score. Considering the results reported for LC and HC datasets, the optimal values of $\lambda_1$ and $\lambda_2$ are $50^{-4}$ and $30^{-2}$, respectively.
    }
    \label{fig:parameters}
\end{figure}

To select the hyperparameters $\lambda_1$ and $\lambda_2$, we use $FID_{output}$ score to rank models trained on 1000 training steps with $\lambda_1, \lambda_2 \in \{10^{-4}, 30^{-4}, 50^{-4}, 10^{-3}, 30^{-3}, 50^{-3}, 10^{-2}, 30^{-2}, 50^{-2}\}$. As observed in Figure~\ref{fig:parameters}, $\lambda_1$ and $\lambda_2$ are set to $50^{-4}$ and $30^{-2}$. To select the best $\lambda_1$ value, we train our model using only the loss action $\mathcal{L}_{action}$. Similarly, the best $\lambda_2$ value is selected by including only the regularization loss $\mathcal{L}_{reg}$ during the training.

\subsection{Ablation tests} \label{sec:ablation}
The ablation results presented in Table~\ref{tab:ablation} demonstrate that omitting the action loss $\mathcal{L}_{action}$ has the largest negative effect on our model's architecture, followed by unfreezing the cross-attention layers during U-Net training. The omission of the regularization term $\mathcal{L}_{reg}$ also negatively impacts performance. The negative effect of omitting the regularization term is more pronounced for the HC dataset than for the LC dataset. This indicates that the regularization loss is particularly important when the model is trained to adapt the environment of the input image without allowing hallucinations, as in the case of the HC dataset.

\subsection{Computational time} \label{sec:time}

While the baselines rely on training-free approaches, the proposed EditAction model requires approximately 9.5 hours of training. However, as demonstrated earlier, EditAction is capable of accurately implementing actions involving both background and foreground objects depicted in the input image while preserving their properties and the background. In contrast, the baselines either produce new images that deviate significantly from the input images or fail to interpret the action-based text instruction, resulting in output images identical to the input images. Therefore, we consider the training time of EditAction to be a justified trade-off for achieving accurate action-based image editing. During the inference, EditAction is as fast as InstructPix2Pix and much faster than the other baselines (Table~\ref{tab:time}).

\begin{table}[h]
\begin{center}
{
\begin{tabular}{l c }
\hline
{Method} & {Time (s)} \\
\hline
InstructPix2Pix &   9.43 \\
 Plug-and-Play & 270.44 \\
 DDPM Noise Space & 85.12 \\
 Prox-MasaCTRL & 18.85 \\
 MasaCTRL & 17.45 \\
EditAction &  9.44 \\
 \hline
\end{tabular}   }
\end{center}
\caption{Computational time during the inference for EditAction and baselines.}\label{tab:time}
\end{table}

\begin{table*}[t!]
    \centering
    \footnotesize
    \begin{tabular}{ll ccc}
    \toprule
      \text{Dataset} & \text{Methods} & $\text{$Acc$}(\uparrow$) &  $\text{$FID_{input}$}(\downarrow$)  &  $\text{$FID_{output}$}(\downarrow$)   \\

      \midrule
      $LC_{test500}$ & w/o $\mathcal{L}_{action}$ & 69.53 $\pm$ 0.24 & 55.78 $\pm$ 0.77  & 58.14 $\pm$ 0.81 \\
      & w/o $\mathcal{L}_{reg}$ & 70.97 $\pm$ 0.11 & 54.78 $\pm$ 0.89 & 57.02 $\pm$ 0.95 \\
      & w/o freezing U-Net & 70.25 $\pm$ 0.28 & 55.43 $\pm$ 0.94 & 57.57 $\pm$ 0.96 \\
    \midrule
    & EditAction & \textbf{71.40 $\pm$ 0.41} & \textbf{54.54 $\pm$ 0.91} & \textbf{56.18 $\pm$ 0.88} \\

      \midrule
      $HC_{test500}$ & w/o $\mathcal{L}_{action}$ & 104.68 $\pm$ 0.22 & 0.99 $\pm$ 1.23 & 104.34 $\pm$ 1.23 \\
      & w/o $\mathcal{L}_{reg}$ & 53.92 $\pm$ 17 & 98.32 $\pm$ 0.97 & 99.11 $\pm$ 0.91 \\
      & w/o freezing U-Net & 53.12 $\pm$ 0.36  & 101.86  $\pm$ 1.0 & 103.23 $\pm$ 1.05 \\
    \midrule
    & EditAction & \textbf{54.41 $\pm$ 0.49} & \textbf{95.76 $\pm$ 0.93}  & \textbf{97.18 $\pm$ 0.82} \\
    
    \bottomrule
    \end{tabular}
    \caption{Ablation results presented for the $LC_{test500}$ and $HC_{test500}$ datasets and evaluated in terms of accuracy, $FID_{input}$ and $FID_{output}$ (mean and variance).}\label{tab:ablation}
\end{table*}

\begin{figure*}[t]
    \centering
    \includegraphics[width=16cm]{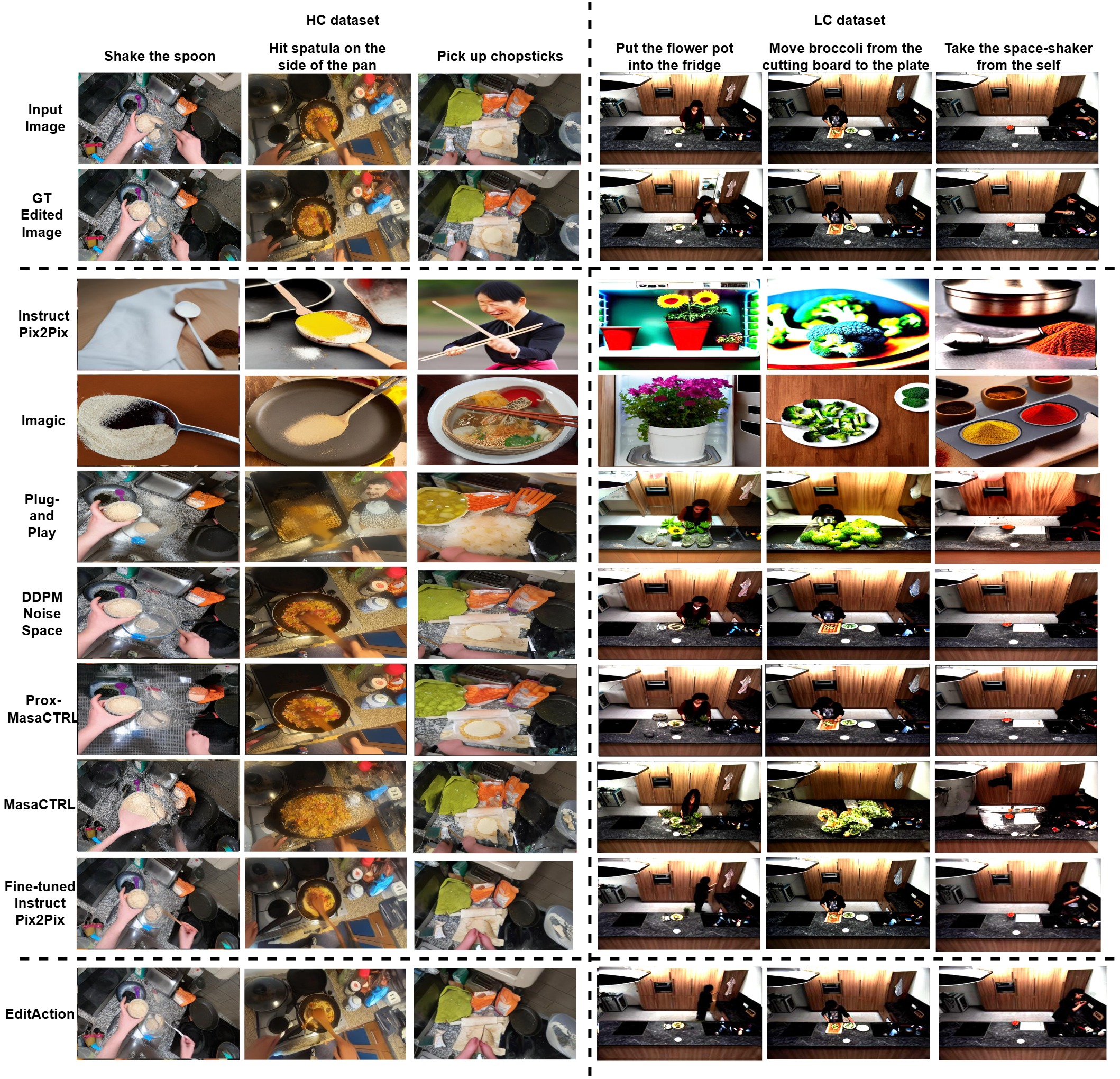}
    \caption{Comparison between EditAction and the baselines on LC and HC datasets.
    }
    \label{fig:appendix_main_figure}
\end{figure*}


\end{document}

%% file: preamble.tex
\usepackage[dvipsnames]{xcolor}


%% file: sec/0_abstract.tex
\begin{abstract}

Text-based image editing is typically approached as a static task that involves operations such as inserting, deleting, or modifying elements of an input image based on human instructions. Given the static nature of this task, in this paper, we aim to make this task dynamic by incorporating actions. By doing this, we intend to modify the positions or postures of objects in the image to depict different actions while maintaining the visual properties of the objects. To implement this challenging task, we propose a new model that is sensitive to action text instructions by learning to recognize contrastive action discrepancies. The model training is done on new datasets defined by extracting frames from videos that show the visual scenes before and after an action. We show substantial improvements in image editing using action-based text instructions and high reasoning capabilities that allow our model to use the input image as a starting scene for an action while generating a new image that shows the final scene of the action.

\end{abstract}

%% file: sec/1_intro.tex
\section{Introduction}
\label{sec:intro}

Image editing based on human instructions has gained considerable attention recently \cite{DBLP:conf/cvpr/BrooksHE23, DBLP:conf/cvpr/KawarZLTCDMI23, DBLP:conf/nips/Li0H23, DBLP:conf/iccv/ZhangRA23, DBLP:conf/iclr/HertzMTAPC23}, especially since the debut of diffusion models. Due to their high scalability, diffusion models are easily adaptable to different domains and support training-free approaches \cite{DBLP:conf/cvpr/KawarZLTCDMI23, DBLP:conf/iccv/ZhangRA23, DBLP:conf/iclr/HertzMTAPC23}. Although numerous models have been proposed to guide image editing using written instructions, most are successful only for static editing, focusing on replacing, inserting, or removing objects or people from an input image \cite{DBLP:journals/corr/abs-2303-04671, DBLP:conf/icml/ChangZBML00MFRL23, DBLP:conf/cvpr/SheyninPSKZAPT24, DBLP:conf/nips/0060YYB023, DBLP:journals/corr/abs-2404-18020, DBLP:conf/cvpr/HuangXWYCG00HZS24, DBLP:conf/siggraph/MaL0L24, DBLP:conf/nips/Li0H23, DBLP:conf/icml/0006YMXE024, DBLP:conf/iccv/ZhangRA23, DBLP:conf/iclr/HertzMTAPC23} or style manipulation and personalization of objects \cite{DBLP:journals/tog/GalAABCC23, DBLP:conf/iccv/YangHY23, DBLP:conf/eccv/Bar-TalOFKD22, DBLP:conf/nips/Wang0LSHCWZ23}. Given the static editing capabilities of the current leading models, they are generally unable to depict the action of objects or people within the scene shown in the input image. By implementing actions, the expected edited image should show the objects or people accomplishing the actions specified in the written instructions while retaining their properties from the input image. These properties can refer to features such as colors, textures, facial features, sizes, shapes, or proportions. By preserving these properties, we ensure that both input and edited images refer to the same objects or people.

In this paper, we take a step further by addressing image manipulation as a dynamic task capable of implementing actions described in text instructions. To display actions in an image, we consider two scenarios with varying levels of difficulty. In the simpler scenario, we imagine that the input and the edited images are captured by a fixed camera. Here, the edited image is expected to share the same environment as the input image while depicting an object or person from the input image performing an action. The second scenario is more challenging, as it assumes the camera is flexible and can capture scenes while moving. In this case, we expect the edited image not only to implement the action but also to slightly modify the environment of the input image. The two scenarios highlight the practicality of image editing models guided by action-based text instructions. While these models can be used for creating new virtual worlds, they can also assist agents in navigating and performing various tasks by using the edited image as a target to assess how well the tasks are fulfilled.

In this paper, we approach image editing guided by action-based text instructions as a task similar to video generation. However, instead of generating every consecutive frame of a video, we produce a single frame that depicts the moment when the action described in the text instruction is fulfilled. This setup makes our task more complex than video generation. Instead of relying on multiple consecutive frames, as in video generation, we need to develop a model capable of reasoning about the new positions or postures of objects or people to accurately depict the required action in the edited image.

To realize the task of action-based image editing via text commands, we introduce a novel model that learns to implement the actions of objects and people in an input image by contrasting correct and randomly selected action-focused text instructions. To achieve this, we adapt the conditional generation capability of a diffusion model to work as a classifier that maximizes the implementation of an action at the cost of suppressing the implementation of other randomly selected actions. To mitigate potential deterioration of the image quality or artifacts that might result from training, we impose several regularization strategies. As we aim to enable our model to focus exclusively on editing the action while preserving the environment and the features of the objects and people in the input image, we build our model on top of InstructPix2Pix~\cite{DBLP:conf/cvpr/BrooksHE23}, a state-of-the-art model already trained for static text-based image editing. We further dubbed our model EditAction, given its capability to edit the actions of objects or people displayed in an image. Some images edited by EditAction are shown in Figure~\ref{fig:first_figure}.


Since datasets for action image editing based on human instructions do not exist, we create new datasets for fine-tuning. To accomplish this, we draw inspiration from datasets used for action recognition in videos. We define our datasets by extracting the first and last frames of an action labeled in a video. The first frame serves as the input image, while the last frame corresponds to the edited image.

Our key contributions are:
\begin{enumerate}
    \item We propose an effective model for image editing using action-based text instructions by training a diffusion-based model for static image editing to recognize contrastive action discrepancies. Unlike our proposed EditAction model, other models claiming action-editing capabilities based on text instructions often either generate entirely new images that differ from the input or replicate the input image without applying the required actions.
    \item We demonstrate both quantitatively and qualitatively that our model effectively implements actions while preserving the environment and the appearance of the objects involved in the actions. Additionally, our model exhibits strong reasoning capabilities for expanding the scene of the input image according to the action text command.
    \item We propose two new datasets for action-based image editing using written instructions, containing pairs of scenes depicting moments before and after an action occurs.
\end{enumerate}


\section{Related Works}\label{sec:related_work}

Inducing motion guidance for subjects in images has been widely explored in the creation of 3D scenes that depict the gradual evolution of objects in space during an action. The motion conditionalities range from trajectories \cite{DBLP:conf/cvpr/WangXXL021, DBLP:conf/iccv/Zhang0BP021}  and poses \cite{DBLP:journals/corr/abs-2210-15134} to text-based inputs, either as sentence instructions or action labels \cite{DBLP:conf/cvpr/GuoZZ0JL022, DBLP:conf/eccv/PetrovichBV22, DBLP:conf/eccv/GuoZWC22, DBLP:conf/iccv/KarunratanakulP23, DBLP:conf/iccv/AthanasiouPBV23, DBLP:conf/iccv/PetrovichBV21}. Unlike generating a sequence of object positions that reflect an action over time, usually on a monochromatic background, our model focuses on editing real-world images to display the final position of an object after an action has been completed.

An intuitive approach to implementing actions in an input image is through a dragging system, as seen in \cite{DBLP:conf/iclr/NieG0ZZL24, DBLP:conf/cvpr/ShiXLPYZTB24, DBLP:conf/iclr/MouWSSZ24}, where the user indicates the starting point (the current position of an object) and the ending point (the expected final location after the action). While this user-friendly approach is effective, it can produce confusing results, particularly for edits involving nested objects. An alternative to the dragging system is action-based text instructions, which can provide the necessary information for editing actions in an image.

Text-based image editing is commonly used for static modifications of the objects depicted in an image. Static editing includes operations such as insertion, replacement, and deletion of objects \cite{DBLP:journals/corr/abs-2303-04671, DBLP:conf/icml/ChangZBML00MFRL23, DBLP:conf/cvpr/SheyninPSKZAPT24, DBLP:conf/nips/0060YYB023, DBLP:journals/corr/abs-2404-18020, DBLP:conf/cvpr/HuangXWYCG00HZS24, DBLP:conf/siggraph/MaL0L24, DBLP:conf/nips/Li0H23, DBLP:conf/icml/0006YMXE024, DBLP:conf/iccv/ZhangRA23, DBLP:conf/iclr/HertzMTAPC23} as well as style manipulation and personalization of objects \cite{DBLP:journals/tog/GalAABCC23, DBLP:conf/iccv/YangHY23, DBLP:conf/eccv/Bar-TalOFKD22, DBLP:conf/nips/Wang0LSHCWZ23}.

Complementary to static editing, dynamic approaches have been proposed to place objects of the input images into different postures based on action-based text instructions. To prevent the introduction of artifacts and ensure preservation of both the background and the objects targeted by the action, Imagic \cite{DBLP:conf/cvpr/KawarZLTCDMI23} optimizes the embeddings of the action-based text instruction to generate images that closely resemble the input image. In \cite{DBLP:conf/cvpr/TumanyanGBD23}, preservation of the background and the targeted objects is achieved by editing the input image using not only text instructions but also features that store layout and spatial information. In \cite{DBLP:conf/prcv/HuangLQC23, DBLP:conf/iccv/CaoWQSQZ23}, the objects involved in the action and the background are preserved by constraining the keys and values of cross-attention in the U-Net diffusion models to rely not only on the action text instruction but also on the latent representation of the input image computed using DDIM Inversion \cite{DBLP:conf/iclr/SongME21}. The inversion of the input image is further enhanced in \cite{DBLP:conf/wacv/Han0CZSRGSH0LZJ24, DBLP:conf/cvpr/Huberman-Spiegelglas24}, where both methods demonstrate that robust latent representations of the input image generated by DDIM Inversion or DDPM Inversion \cite{DBLP:conf/nips/HoJA20} improve image structure preservation while accurately implementing actions depicted in the text instructions. While the DDPM Noise Space \cite{DBLP:conf/cvpr/Huberman-Spiegelglas24} is built on top of Stable Diffusion \cite{DBLP:conf/cvpr/RombachBLEO22}, ProxMasaCtrl \cite{DBLP:conf/wacv/Han0CZSRGSH0LZJ24} extends MasaCtrl \cite{DBLP:conf/iccv/CaoWQSQZ23}. 

While the above models focus primarily on preserving the background and the objects involved in the actions, the self-guidance diffusion model \cite{DBLP:conf/nips/EpsteinJPEH23} explicitly addresses action implementation by using self- and cross-attention maps, computed by U-Net diffusion models for pointing to the region where the target object is initially located and to the region where it should move after the action.

Unlike the above models, the proposed EditAction model explicitly addresses both the challenges of action implementation and preservation of the background and targeted objects. Additionally, our model demonstrates reasoning capabilities, allowing it to expand the scene of the input image based on the specified action task. A concurrent approach to ours is presented in \cite{DBLP:journals/corr/abs-2408-05802}, where a diffusion-based model is trained to depict actions in input images using optical flows that capture object motion between frames. 

\begin{figure*}[t]
    \centering
    \includegraphics[width=15cm]{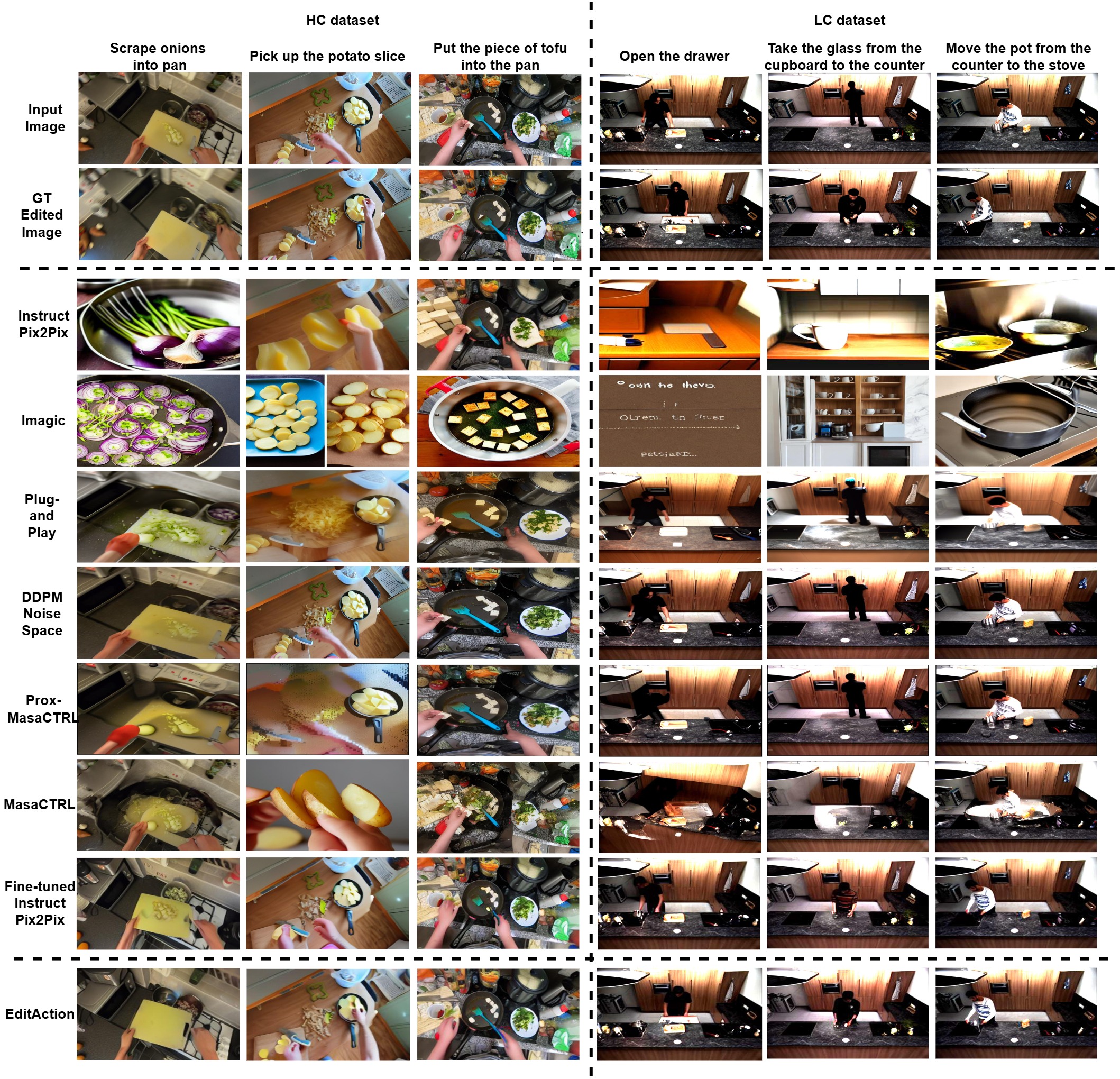}
    \caption{Comparison between EditAction and the baselines on the LC and HC datasets. Unlike the baselines, our model can implement actions while preserving the background and the appearance of the objects involved in the action. More examples are presented in Figure~\ref{fig:appendix_main_figure} (in Appendix).
    }
    \label{fig:main_figure}
\end{figure*}

%% file: sec/2_formatting.tex
\section{Methodology}
\label{sec:method}

To enhance sensitivity toward action recognition, we approach text-based image editing as a supervised task, trained on paired images that depict scenes before and after an action. The preliminaries required to develop our model are presented in Subsection~\ref{sec:Preliminaries}. The datasets used for training are introduced in Subsection~\ref{sec:data} and our model, EditAction, is introduced in Subsection~\ref{sec:editaction}.

\subsection{Preliminaries}\label{sec:Preliminaries}

We based our model for action editing using human instructions on InstructPix2Pix, a diffusion-based model trained for static image editing using a pretrained Stable Diffusion checkpoint. We hypothesize that using a model already trained for noun-based image editing, which performs operations like inserting, removing, or replacing objects or people in an image, will ease our fine-tuning process, as this allows our model to focus solely on learning how to display actions.

The diffusion process implemented by InstructPix2Pix gradually adds noise on the edited image (output image) $x_0$ over $T$ iterations, resulting in the noisy latent $x_T$. Considering the constraints given by the input image $i$ and the text command $c$, during the denoising process defined as a Markov chain, the conditional probability computed by the diffusion process is: 
\begin{equation}\label{eq_1}
    p_{\theta}(x_0|i, c) = \int_{x_{1:T}} p(x_T)\prod_{t=1}^T p_{\theta}(x_{t-1}|x_t, i, c), \,dx_{1:T} \
\end{equation}
where $p(x_T) \sim \mathcal{N}(0, I)$. As $p_{\theta}(x_0|i, c)$ is intractable for complex models, InstructPix2Pix is trained by minimizing the variational lower bound (ELBO) of the log-likelihood $p_{\theta}(x_0|i, c)$. Following the simplified ELBO defined in \cite{DBLP:conf/nips/HoJA20}, InstructPix2Pix learns a U-Net network $\epsilon_\theta$ to predict the noise added to the noisy latent $x_t$:
\begin{multline}\label{eq_2}
\mathcal{L}_{static}(x_0, i, c, \theta) = \log p_{\theta}(x_0|i, c) := \\
\mathbb{E}_{\epsilon, x_0, t} \left[ ||\epsilon - \epsilon_{\theta}(x_t, t, i, c)||_2^2\right] 
\end{multline}
\begin{equation}\label{eq_2_1}
    x_t = \sqrt{\alpha_t}x_0+\sqrt{1-\alpha_t}\epsilon
\end{equation}
where $\epsilon \sim \mathcal{N}(0, \mathcal{I})$ and $\alpha_t = \prod_{i=0}^t{(1-\beta_i)}$ ($\beta_i$ is a coefficient between 0 and 1 that guides the noise level). 

To support the image conditionality, we follow the methodology of InstructPix2Pix and concatenate the noisy latent $x_t$ with the input image $i$. The network  $\epsilon_{\theta}$ is further trained on this concatenated input. The further adaptation of InstructPix2Pix for action-based image editing is presented in Subsection~\ref{sec:editaction}.

\subsection{Datasets for action-based image editing with written instructions}\label{sec:data}

To extract image pairs that show scenes before and after an activity, we rely on datasets designed for action recognition in videos. Given a time interval corresponding to a video segment depicting an activity, we select the first frame of the segment as the input image (before the action) and the last frame as the edited image (after the action). Using this approach, we choose datasets for action recognition in videos that involve actions with clear starting and ending points in space. Additionally, the selected datasets must have well-defined frame boundaries to accurately identify the starting and ending scenes of an action.

The low-complexity (LC) dataset is built using a fixed camera, ensuring that the input and edited images always share the same environment. This dataset is based on MPII-Cooking~\cite{DBLP:journals/ijcv/RohrbachRRAAPS16}, a collection of videos recording people preparing various dishes. Each action-based text instruction contains a verb depicting the activity and the object targeted by the action. Additionally, the instructions may include spatial details indicating the starting and ending points of the activity. Unlike the LC dataset, the high-complexity (HC) dataset contains images extracted from videos recorded with a flexible camera. To define the HC dataset, we use EPIC-Kitchen~\cite{DBLP:journals/ijcv/DamenDFFKMMMPPW22}, a video collection of domestic activities. The higher complexity of the HC dataset, compared to the LC dataset, arises not only from the flexible camera (head camera) but also from the text instructions. In the HC dataset, the text instructions are less informative than those in the LC dataset, usually specifying only the action verb and the object. Some instances of the two datasets are presented in Figure~\ref{fig:datasets}. More details about the LC and HC datasets are presented in Appendix~\ref{sec:datasets}.

\begin{figure}[t]
    \centering
    \includegraphics[width=8.5cm]{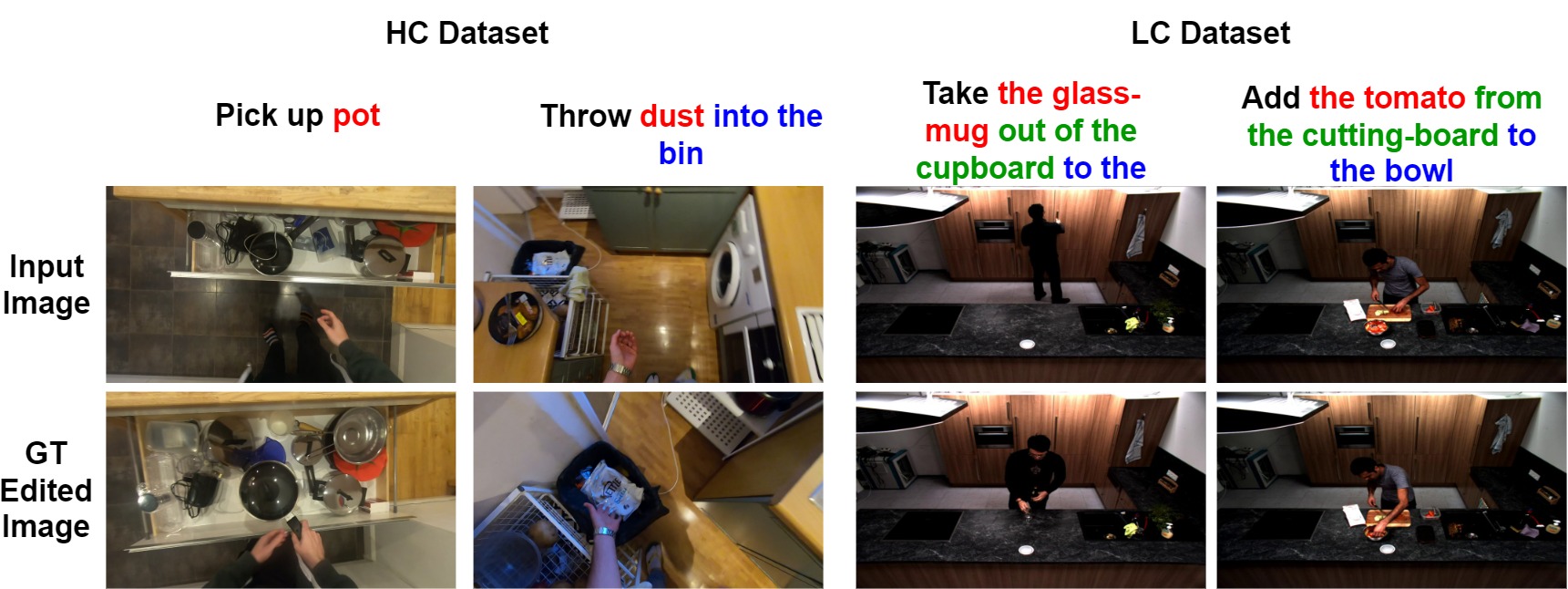}
    \caption{Left side: HC dataset; Right side: LC dataset. \textcolor{red}{Red} indicates the object targeted by the action. \textcolor{green}{Green} and \textcolor{blue}{blue} shows the starting and the ending points of the action.
    }
    \label{fig:datasets}
\end{figure}

\begin{figure}[t]
    \centering
    \includegraphics[width=8.5cm]{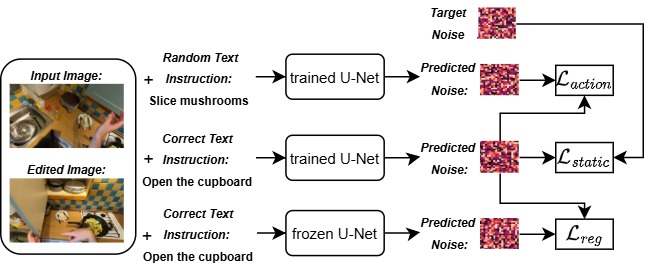}
    \caption{Training of the U-Net model employed by EditAction. Given an input and an edited image, the model is trained to enhance alignment with the action-based text instruction while preventing hallucinations that may result from fine-tuning on small-scale datasets.}
    
    \label{fig:overview}
\end{figure}

\subsection{EditAction}\label{sec:editaction}

To implement an activity, we adapt the diffusion-based model to contrastively differentiate between the text instruction $c$ that specifies the activity and another randomly selected action-based text instruction $c{'}$ that does not share any token with $c$. Given the output image $x_0$, we train our model to maximise the likelihood of predicting $c$, $p_{\theta}(c|x_0, i)$, while reducing the probability of predicting a random text instruction.
To compute $p_{\theta}(c|x_0, i)$, we need to adapt the diffusion model developed for conditional generation to work as a classifier. As suggested in \cite{DBLP:conf/iccv/LiPDBP23}, the conversion from conditional generation to classification can be done following Bayes' theorem. Given the text instruction $c$ and the randomly selected text command $c{'}$, $p_{\theta}(c|x_0, i)$ is computed as:
\begin{equation}\label{eq_3}
    \frac{p(c)p_{\theta}(x_0|i, c)}{\sum_{c_j \in (c, c{'})}  p(c_j)p_{\theta}(x_0|i, c_j)}
\end{equation}

Assuming a uniform distribution over the text commands $c_j$ ($p(c_j) = \frac{1}{2}$, as $c_j \in (c, c^{'}))$, all the priors $p(c_j)$ are removed. To estimate $p_{\theta}(x_0|c, i)$, an intractable likelihood, we rely again on the ELBO estimation of $\log p_{\theta}(x_0|c, i)$ given in Eq. \ref{eq_2}. In the end, to maximise the likelihood $p_{\theta}(c|x_0, i)$, we train our model by minimizing the following loss:
\begin{multline}
    \mathcal{L}_{action}(x_0, i, c, \theta) = \log p_{\theta}(c|x_0, i) := \\
    \frac {\exp \{ \mathbb{E}_{\epsilon, x_0, t} \left[ ||\epsilon - \epsilon_{\theta}(x_t, t, i, c)||_2^2\right] \}  }
    {\sum_{c_j \in (c, c{'})} \exp \{ \mathbb{E}_{\epsilon, x_0, t} \left[ ||\epsilon - \epsilon_{\theta}(x_t, t, i, c_j)||_2^2\right] \}}
\end{multline}

Fine-tuning the model on new data risks discounting the knowledge gained during the previous training (Stable Diffusion and InstructPix2Pix training phases). This risk can result in overfitting, hallucination and deterioration of image quality. However, as demonstrated by Fan et. al~\cite{DBLP:journals/corr/abs-2305-16381} adding a new regularization term in the fine-tuning process can prevent this. We apply the same concept to our model and aim to prevent disruptions of the data distribution learnt by $\epsilon_{\theta}$ during fine-tuning and training. Knowing that $\epsilon_{\theta}^{freeze}$ represents the U-Net network before fine-tuning, while $\epsilon_{\theta}$ is the network that we fine-tune, we define the regularization term as follows:
\begin{multline}
    \mathcal{L}_{reg}(x_0, i, c, \theta) = \\
    \mathbb{E}_{x_0, t} \left[ ||\epsilon_{\theta}^{freeze}(x_t, t, i, c) - \epsilon_{\theta}(x_t, t, i, c)||_2^2\right]
\end{multline}

To further reduce the loss of training knowledge that can affect the capability of static image editing inherited from InstructPix2Pix, we freeze all cross-attention layers of the network $\epsilon_{\theta}$ that connect the text and visual modalities. While the cross-attention layers are frozen, we fine-tune the remaining layers. As InstructPix2Pix is trained to edit images statically, we hypothesise that freezing the cross-attention layers will conserve the alignments between the text and visual representations, and EditAction will be able to show in the edited image the objects and their visual properties displayed in the input image. Another advantage of freezing the cross-attention layers is reducing the GPU memory required for training by up to 45\%.

Finally, all the loss functions are integrated and the fine-tuning of our model is done as follows:
\begin{multline}
    \mathcal{L}(x_0, i, c, \theta) = \mathcal{L}_{static}(x_0, i, c, \theta) + \\
    \lambda_1 \mathcal{L}_{action}(x_0, i, c, \theta) + \lambda_2 \mathcal{L}_{reg}(x_0, i, c, \theta)
\end{multline}
where $\lambda_1$ and $\lambda_2$ are the weights of the action and regularization losses. The training procedure is illustrated in Figure~\ref{fig:overview}.

During the inference, the integration of the visual and text modalities represented by the input image $i$ and a text command $c$ into the fine-tuned model $\epsilon_{\theta}$ is done as follows~\cite{DBLP:conf/cvpr/BrooksHE23}:
\begin{multline}
    \hat {\epsilon_{\theta}}(x_t, t, i, c) = \epsilon_{\theta}(x_t, t) + \\
    s_i \cdot (\epsilon_{\theta}(x_t, t, i) - \epsilon_{\theta}(x_t, t)) \\
    s_c \cdot (\epsilon_{\theta}(x_t, t, i, c) - \epsilon_{\theta}(x_t, t, i))
\end{multline}
where $s_i$ and $s_c$ represent the guidance scales for image and text conditionalities.

\section{Experimental Set-up}\label{sec:setup}

\subsection{Datasets and models}\label{datasets}
\textbf{Datasets.} To define the test datasets, we randomly selected 500 image pairs from the LC and HC datasets, dubbing them $LC_{test500}$ and $HC_{test500}$. Additionally, we manually selected 100 image pairs from the HC dataset, referred to as $HC_{test100}$, which depict \textit{long-distance} actions where the central visual field in the input image differs from the central visual field of the edited image (as described by the text instruction "throw dust into the bin" in Figure~\ref{fig:datasets}). We trained EditAction on the remaining instances, which include 13266 instances from the LC dataset and 36854 instances from the HC dataset.

\begin{table*}[t!]
    \centering
    \footnotesize
    \begin{tabular}{l | ccc |ccc}
    \toprule
      & \multicolumn{3}{c}{$LC_{test500}$}  &  \multicolumn{3}{c}{$HC_{test500}$ }  \\
      \midrule
       Method & \multicolumn{1}{c}{$\text{$Acc$}(\uparrow$)}  &  \multicolumn{1}{c}{$\text{$FID_{input}$}(\downarrow$) } 
       &  \multicolumn{1}{c}{$\text{$FID_{output}$}(\downarrow$) } & \multicolumn{1}{c}{$\text{$Acc$}(\uparrow$)} &  \multicolumn{1}{c}{$\text{$FID_{input}$}(\downarrow$) } &  \multicolumn{1}{c}{$\text{$FID_{output}$}(\downarrow$) }  \\
      \midrule
\text{Imagic} & 12.81 $\pm$ 0.32 & 356.26 $\pm$ 0.97 & 356.19 $\pm$ 1.02 & 29.46 $\pm$ 0.22 & 206.18 $\pm$ 0.74 & 205.61 $\pm$ 1.22 \\
\text{Plug-and-Play} & 48.05 $\pm$ 0.43 & 92.27 $\pm$ 0.67 & 93.57 $\pm$ 0.65 & 44.84 $\pm$ 0.14 & 123.08 $\pm$ 1.02 & 123.23 $\pm$ 0.65 \\
\text{DDPM Noise Space} & 65.26 $\pm$ 0.22 & 60.99 $\pm$ 0.72 & 63.6 $\pm$ 0.77 & 46.27 $\pm$ 0.21 & \textbf{93.9 $\pm$ 0.95} & 105.73 $\pm$ 0.78 \\
\text{Prox-MasaCTRL} & 64.64 $\pm$ 0.53 & 83.87 $\pm$ 0.74 & 85.42 $\pm$ 0.92 & 46.25 $\pm$ 0.36 & 118.82 $\pm$ 0.82 & 121.23 $\pm$ 0.83 \\
\text{MasaCTRL} & 47.49 $\pm$ 0.52 & 109.57 $\pm$ 0.85 & 108.24 $\pm$ 0.99 & 50.28 $\pm$ 0.25 & 152.1 $\pm$ 0.85 & 158.49 $\pm$ 0.77 \\
\text{InstructPix2Pix} & 21.29 $\pm$ 0.37 & 270.49 $\pm$ 0.78 & 270.97 $\pm$ 0.73 & 32.86 $\pm$ 0.32 & 185.29 $\pm$ 0.79 & 186.25 $\pm$ 0.59 \\
\text{Fine-tuned InstructPix2Pix} & 69.2 $\pm$ 0.33 & 55.88 $\pm$ 0.94 & 60.12 $\pm$ 0.82 & 52.84 $\pm$ 0.33 & 105.97 $\pm$ 0.73 & 99.01 $\pm$ 0.61 \\
\midrule
\text{EditAction} & \textbf{71.4 $\pm$ 0.41} & \textbf{54.54 $\pm$ 0.91} & \textbf{56.18 $\pm$ 0.88} & \textbf{54.41 $\pm$ 0.49} & 95.76 $\pm$ 0.93 & \textbf{97.28 $\pm$ 0.82} \\
    \bottomrule
    \end{tabular}
\caption{Quantitative evaluation of EditAction compared to baseline models for the $LC_{test500}$ and $HC_{test500}$ datasets. The evaluation relies on the accuracy of action recognition and the FID score computed with respect to the input image ($FID_{input}$) and the ground-truth edited image ($FID_{output}$) (mean and variance).}\label{tab:quantitative_results}
\end{table*}

\begin{table*}[t!]
    \centering
    \footnotesize
    \begin{tabular}{l |cccc |cccc |c}
    \toprule
      & \multicolumn{4}{c}{$LC_{test500}$}  &  \multicolumn{4}{c}{$HC_{test500}$} & $HC_{test100}$ \\
      \midrule
       Method & \multicolumn{1}{c}{$\text{$R1$}(\uparrow$)}  &  \multicolumn{1}{c}{$\text{$R2$}(\uparrow$)}  & \multicolumn{1}{c}{$\text{$R3$}(\uparrow$)}  & \multicolumn{1}{c}{$\text{$R4$}(\uparrow$)}  & \multicolumn{1}{c}{$\text{$R1$}(\uparrow$)}  &  \multicolumn{1}{c}{$\text{$R2$}(\uparrow$)}  & \multicolumn{1}{c}{$\text{$R3$}(\uparrow$)}  & \multicolumn{1}{c}{$\text{$R4$}(\uparrow$)}  &
       \multicolumn{1}{c}{$\text{$R5$}(\uparrow$)}\\
      \midrule
\text{Imagic} & 1.56 & 1.78 & 1.54 & 1.33 & 1.63 & 1.43 & 1.47 & 1.32 & 1.23 \\ 
\text{Plug-and-Play} & 3.28 & 3.78 & 4.22 & 2.67 & 3.25 & 3.71 & 3.84 & 2.88 & 2.67 \\ 
\text{DDPM Noise Space} & 3.31 & 4.31 & \textbf{4.75} & 2.88 & 3.25 & 4.25 & \textbf{4.64} & 2.98 & 3.12 \\ 
\text{Prox-MasaCTRL} & 3.32 & 4.29 & 4.71 & 2.79 & 3.28 & 4.13 & 4.48 & 2.93 & 3.17 \\ 
\text{MasaCtrl} & 3.01 & 3.43 & 3.62 & 2.74 & 3.62 & 3.45 & 3.49 & 3.12 & 3.39 \\ 
\text{InstructPix2Pix} & 1.68 & 1.74 & 1.76 & 1.46 & 1.73 & 2.19 & 2.33 & 1.93 & 1.57 \\ 
\text{Fine-tuned instruct Pix2Pix} & 4.23 & 4.48 & 4.45 & 4.28 & 4.25 & 4.56 & 4.57 & 4.31 & 3.9 \\ 
\midrule
\text{EditAction} & \textbf{4.37} & \textbf{4.68} & 4.59 & \textbf{4.53} & \textbf{4.28} & \textbf{4.61} & 4.57 & \textbf{4.44} & \textbf{3.98} \\

    \bottomrule
    \end{tabular}
    \caption{Qualitative evaluation of EditAction compared to the baseline models for the $LC_{test500}$, $HC_{test500}$ and $HC_{test100}$ datasets.}\label{tab:qualitative_results}
\end{table*}

\noindent\textbf{Models.} We compare our model EditAction with Imagic \cite{DBLP:conf/cvpr/KawarZLTCDMI23}, Plug-and-Play \cite{DBLP:conf/cvpr/TumanyanGBD23}, DDIM Inversion \cite{DBLP:conf/cvpr/Huberman-Spiegelglas24}, Prox-MasaCTRL \cite{DBLP:conf/wacv/Han0CZSRGSH0LZJ24} and MasaCTRL \cite{DBLP:conf/iccv/CaoWQSQZ23}. Additional baselines are InstructPix2Pix \cite{DBLP:conf/cvpr/BrooksHE23} and its version fine-tuned on LC and HC datasets. The other image editors for action implementation and discussed above \cite{DBLP:conf/prcv/HuangLQC23, DBLP:conf/nips/EpsteinJPEH23, DBLP:journals/corr/abs-2408-05802} are not evaluated due to the lack of available source code.

\subsection{Metrics}\label{metrics}

\textbf{Quantitative evaluation. }The results are quantitatively assessed using the Fréchet Inception Distance (FID) \cite{Heusel2017FID}. We use FID to measure the similarity between the generated image and the input image, as well as the similarity between the generated image and the ground-truth edited image. Although the CLIP score \cite{Hessel2021CLIPscore} is a commonly used metric for evaluating text-image similarity, we do not use it here. Our text instructions do not describe the entire image, as required by the CLIP score, but instead provide action indications for objects that may not appear in the image's foreground. As an alternative for the CLIP Score, we use TimeSformer \cite{DBLP:conf/icml/BertasiusWT21} to recognise the actions depicted between input and the edited images. 

To identify the actions, we fine-tune TimeSformer separately on videos from the MPII-Cooking and EPIC-Kitchens datasets, which are used to define the LC and HC datasets. Given an input and a ground-truth edited image, we select the video segment where the input image corresponds to the first frame and the ground-truth edited image to the last frame. For the LC dataset, we fine-tune TimeSformer using the checkpoint trained on the Something-Something-V2 dataset \cite{DBLP:conf/iccv/GoyalKMMWKHFYMH17}. For the HC dataset, TimeSformer is trained on the checkpoint obtained by pre-training the model on the Kinetics-600 dataset \cite{DBLP:journals/corr/abs-1808-01340}\footnote{We use the implementation details specific to TimeSformer: \url{https://github.com/facebookresearch/TimeSformer}.}. Since TimeSformer is trained to recognize actions in videos using 8 video frames as input, during inference, we replicate the input image 4 times and the edited image another 4 times. This process is done using both generated and ground-truth edited images. The matching between labels is evaluated based on accuracy (Acc). 

\begin{table}[t!]
    \centering
    \footnotesize
    \begin{tabular}{l | cc}
    \toprule
       Method & \multicolumn{1}{c}{$\text{$Acc$}(\uparrow$)}  &  \multicolumn{1}{c}{$\text{$FID_{output}$}(\downarrow$) }  \\
      \midrule
\text{Imagic} & 28.9 $\pm$ 0.28 & 199.84 $\pm$ 0.86\\
\text{Plug-and-Play} & 38.23 $\pm$ 0.23 & 132.53 $\pm$ 0.76\\
\text{DDPM Noise Space} & 36.54 $\pm$ 0.28 & 112.4 $\pm$ 0.74\\
\text{Prox-MasaCTRL} & 35.17 $\pm$ 0.15 & 108.74 $\pm$ 0.88\\
\text{MasaCtrl} & 39.84 $\pm$ 0.19 & 141.41 $\pm$ 0.86\\
\text{InstructPix2Pix} & 33.64 $\pm$ 0.24 & 187.33 $\pm$ 0.82\\
\text{Fine-tuned instruct Pix2Pix} & 45.89 $\pm$ 0.27 & 104.56 $\pm$ 0.95\\
\midrule
\text{EditAction} & \textbf{46.29 $\pm$ 0.29} & \textbf{102.85 $\pm$ 0.94}\\

    \bottomrule
    \end{tabular}
    \caption{Quantitative evaluation of EditAction compared to baseline models on the $HC_{test100}$ dataset. The evaluation is based on accuracy and the FID score computed with respect to the ground-truth edited image ($FID_{output}$) (mean and variance). 
    }\label{tab:quantitative_results_hc_100test}
\end{table}

\textbf{Qualitative evaluation. }For qualitative evaluation, we show to the ClickWorker annotators the ground truth input and edited images alongside the action-based text instruction and the edited images generated by our model and the baselines. The annotators are asked to assess the quality of action-based image editing, the preservation of the environment and the object targeted by action, and the reasoning capabilities of the models. 

To evaluate how well the actions are implemented in the edited images, we ask annotators to assess the following requirements: "Given the input image, evaluate how well the generated edited image reflects the action $\langle$\textit{action}$\rangle$ described in the text instruction, regardless of the ground-truth edited image." (R1), and "Evaluate how similar the generated edited image is to the ground-truth edited image." (R2).

Although action-based image editing is a dynamic task, we still expect the models to preserve the environment and most of the objects. To evaluate environment preservation, annotators are asked the following requirement: "Evaluate how well the generated edited image preserves the environment and the objects observed in the input image." 
(R3). For the HC dataset, annotators are informed: "Consider that the generated edited images should simulate a scene observed by a head camera directed towards the action specified in the text instruction. Therefore, some marginal objects from the input image might not appear in the generated edited image as long as the action is properly implemented." Additionally, we ask annotators: "Score how well the position or posture of the object $\langle$\textit{object}$\rangle$  targeted by the action $\langle$\textit{action}$\rangle$ is implemented while the properties of the object are preserved." (R4). We evaluate the requirements R1-R4 for both $LC_{test500}$ and $HC_{test500}$ datasets. To assess reasoning capabilities for the $HC_{test100}$ subset, annotators are asked: "Imagine you are wearing a head camera, and the input image shows what you see before performing the action $\langle$\textit{action}$\rangle$. Evaluate, how well the generated image portrays what you should see after completing the action." (R5).

\subsection{Implementation details}\label{implementation}
To edit image actions using text instructions, we fine-tuned InstructPix2Pix, which was developed on top of the Stable Diffusion v1.5 checkpoint. We trained our model for action editing on an NVIDIA GeForce RTX GPU with 24GB of GPU RAM. The batch size was set to 64, and the number of training steps was set to 10000. Following the training of InstructPix2Pix, we set the learning rate to $10^{-4}$ and the image resolution to 256 $\times$ 256. During inference, the image resolution is increased to 512 $\times$ 512, and the number of denoising steps is set to 100. The hyperparameters $s_i$ and $s_c$ are set to 1 and 7.5, as suggested for InstructPix2Pix. We generate images using three seed values and report the quantitative results using mean and variance. The hyperparameters $\lambda_1$ and $\lambda_2$ are set to $50^{-4}$ and $30^{-2}$. Their selection is discussed in the Appendix \ref{sec:editaction_hyperparamets}

\begin{figure}[t]
    \centering
    \includegraphics[width=6cm]{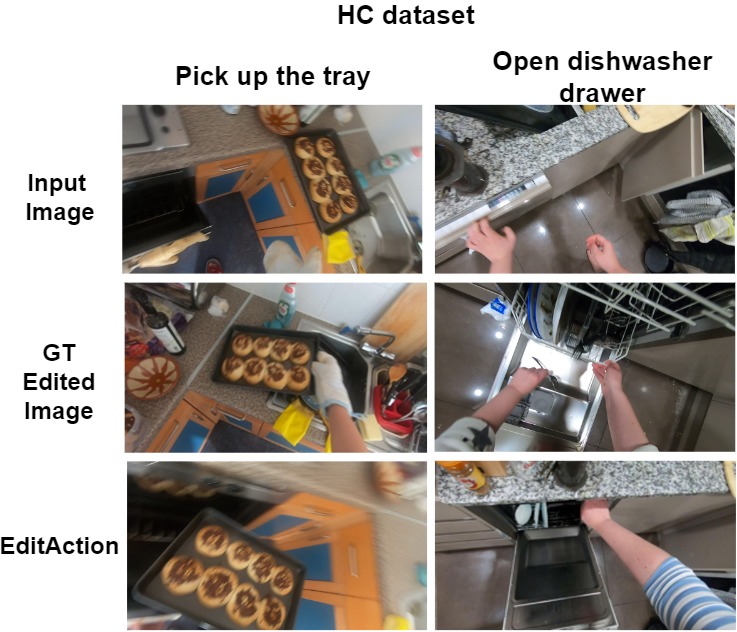}
    \caption{Illustrations of EditAction's reasoning capabilities, which allow it to extend the scene of the input image based on the action-based text command.
    }
    \label{fig:reasoning}
\end{figure}

\begin{figure}[t]
    \centering
    \includegraphics[width=6cm]{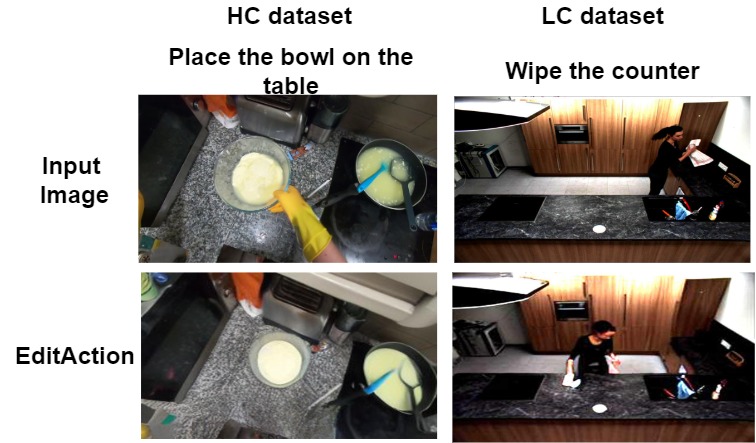}
    \caption{Despite being trained on a limited number of actions, EditAction can implement actions using verbs unseen during the training, such as "wipe" for the LC dataset and "place" for the HC dataset.
    }
    \label{fig:unseen_verbs}
\end{figure}

\begin{figure}[t]
    \centering
    \includegraphics[width=8cm]{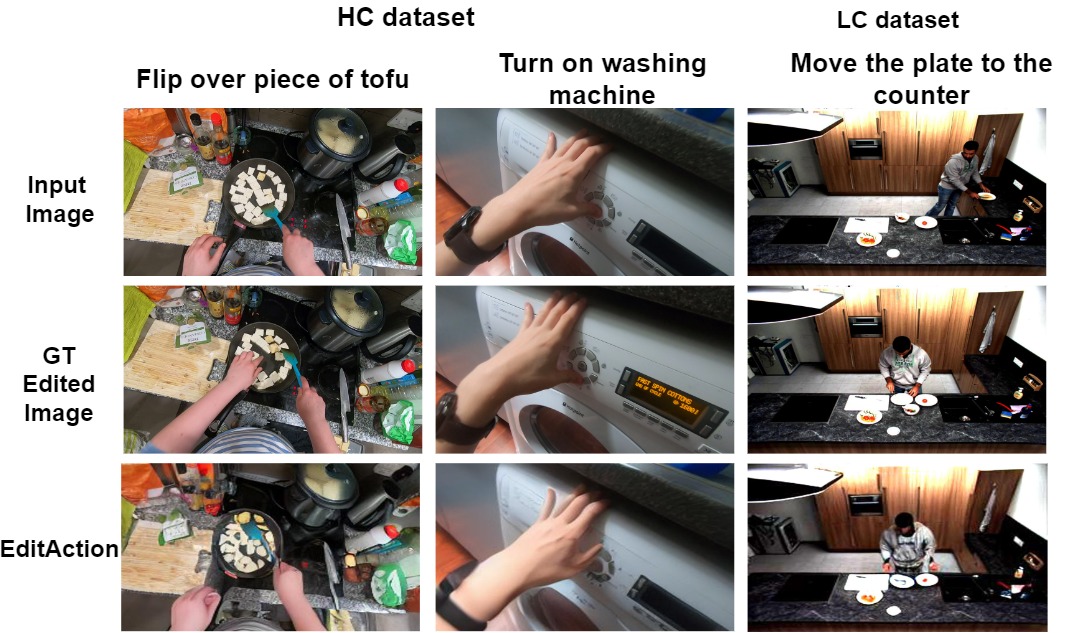}
    \caption{Limitations of EditAction.
    }
    \label{fig:failed_cases}
\end{figure}

\section{Results}\label{sec:results}

This section discusses the quantitative and qualitative evaluation of EditAction. The ablations results are presented in Appendix \ref{sec:ablation}.

\textbf{Quantitative evaluation. }Table \ref{tab:quantitative_results} presents the quantitative results for the $LC_{test500}$ and $HC_{test500}$ test datasets. Compared to the baselines, EditAction consistently better implements the actions described in the text instructions, as demonstrated by the accuracy metric, while preserving both the output and the object targeted by the action, as indicated by the FID score. As expected, EditAction performs more effectively on $LC_{test500}$ than on $HC_{test500}$ due to the fixed camera setup. Among the baselines, we observe that Imagic and InstructPix2Pix have the lowest ability to implement actions. As illustrated in Figure \ref{fig:main_figure}, both Imagic and InstructPix2Pix tend to generate images different from the input, showing only the targeted objects without depicting the actions. While DDPM Noise Space achieves the best results among the baselines for image similarity, its effectiveness is primarily due to its ability to preserve the background without implementing the action. In terms of action implementation, MasaCTRL shows the best performance among the baselines, particularly when the targeted objects are in the foreground, as seen in the HC dataset. However, this comes at the cost of altering both the background and the objects targeted by the action.

To evaluate the reasoning ability of EditAction, we use the $HC_{test100}$ dataset. Since we expect the generated edited images to significantly differ from the input images, the comparison between EditAction and the baselines is not based on $FID_{input}$. As shown in Table \ref{tab:quantitative_results_hc_100test}, the substantial difference between EditAction and the baselines in terms of accuracy and $FID_{output}$ demonstrates our model's ability to reason over the input image and imagine a new scene that reflects the required action. This behavior is illustrated in Figure \ref{fig:reasoning}, where EditAction generates images showing a picked-up tray with a different view of the same kitchen observed in the input image, or an opened dishwasher with a different interior while preserving the rest of the kitchen setup from the input image.

\textbf{Qualitative evaluation. }The results discussed above are confirmed by the annotators in Table \ref{tab:qualitative_results}. As seen for requirements $R1$, $R2$ and $R3$, EditAction can better implement the actions than the baselines while keeping the background. As DDPM Noise Space and Prox-MasaCTRL tend to keep the input image unaltered, they score better than EditAction for requirement $R3$. Regarding requirement $R4$, we notice that EditAction assures better posture and position implementation than the baselines based on the action-based text instructions while properly keeping the properties of the targeted objects. For requirement $R5$, the annotators confirm that while EditAction is less effective at implementing \textit{long-distance} actions specific to the $HC_{test100}$ dataset than the randomly selected actions of the $HC_{test500}$ dataset, it still provides better reasoning and editing capabilities than its baselines.

\textbf{Implementation of unseen actions. }Despite being trained on a limited set of actions, EditAction can implement text instructions containing verbs not encountered during training. Such examples are presented in Figure \ref{fig:unseen_verbs}, where EditAction can implement new actions given by verbs like ``place" for the HC dataset or ``wipe" for the LC dataset.


\textbf{Failed cases. }Although effective for editing actions in input images, EditAction can become confused when the action refers to an object while the input image already contains other similar objects. As shown in Figure \ref{fig:failed_cases}, the model may fail to flip the correct piece of tofu or forget to move the plate to the counter near the other two existing plates. Additionally, EditAction may exhibit limited reasoning capabilities, as demonstrated in the second example, where turning on the washing machine only results in changing the position of the finger without displaying a message on the screen.

\textbf{Limitations. }Despite our model’s effectiveness, a primary limitation is its training requirement. While we show that training is important for implementing actions of small objects that may not be in the foreground and for enabling reasoning during editing, this training process remains a cost of our approach. Additional details about the training and inference times for our model and the baselines are provided in the Appendix~\ref{sec:time}.

\section{Conclusions}\label{sec:conclusion}



While text-based image editing often focuses on operations like inserting, modifying, replacing, or personalizing objects, our work aims to draw attention to action implementation in image editing. Specifically, we aim to alter the positions or postures of objects (or people) in input images to reflect actions specified by text instructions. To accomplish this, we introduce a novel model that implements actions by learning contrasting action-based discrepancies between correct and randomly selected text instructions. Across two scenarios with varying difficulty, one showing images captured with a fixed camera and the other showing images taken with a flexible camera, we demonstrate our model’s effectiveness in action-oriented image editing. In contrast to our approach, existing text-based image editors claiming action editing capabilities tend to either leave the input image unchanged or generate entirely new images that differ from the input.